# ON THE SPATIAL STRUCTURE OF MIXTURE-OF-EXPERTS IN TRANSFORMERS

**Daniel Bershatsky**
AI Center, Skoltech
Moscow, Russia
`daniel.bershatsky2@skoltech.ru`

**Ivan Oseledets**
AIRI
Moscow, Russia
`oseledets@airi.net`

## ABSTRACT

A common assumption is that MoE routers primarily leverage semantic features for expert selection. However, our study challenges this notion by demonstrating that positional token information also plays a crucial role in routing decisions. Through extensive empirical analysis, we provide evidence supporting this hypothesis, develop a phenomenological explanation of the observed behavior, and discuss practical implications for MoE-based architectures.

## 1 INTRODUCTION

The integration of the Mixture of Experts (MoE) approach, originally proposed by (Jacobs et al., 1991) and (Jordan and Jacobs, 1994), into TRANSFORMER-based models has been a key driver of recent advancements in machine learning, particularly in natural language processing (NLP) (Dai et al., 2024; Muennighoff et al., 2024; Qwen et al., 2024; 2025). This innovation enables models to scale efficiently, achieving higher overall parameter counts and improved performance on downstream tasks while maintaining manageable computational requirements for training.

Mixture of Experts (MoE) approach forms a common building block which includes a set of "experts", typically neural networks of the same architecture but different weights, and a "router", a linear multi-class classifier that selects experts. The construction usually substitutes feed-forward networks in TRANSFORMERS-blocks. Only a limited subset of experts is used for processing a single input (or token), which is an appealing feature of MoE in training and inference.

General consensus is that experts are semantically specialized, and combined with dynamical gating, become an appealing approach for building large language models (LLMs). However, preliminary experimental results show significant role of token position, challenging this assumption. In this work, we address this misconception and provide multiple experiments that support our hypothesis.

**Hypothesis 1.1** *(Main observation)*  Mixture-of-Experts (MoE) router in TRANSFORMERS exploits positional token information in addition to semantic one.

We present experimental evidence supporting our hypothesis in Section 2. Further, we offer a phenomenological explanation and discuss empirical findings in more formal terms in Section 3. Finally, we explore the practical implications of both the empirical observations and the phenomenological model in Section 4.

## 2 EMPIRICAL STUDY

We begin with a preliminary experiment in Section 2.1 in order to gain a general understanding of expert activations. Next, we estimate the correlations between activations of different experts Section 2.2. Additionally, we train a classifier on embedded token sequences to predict token posi-tions in Section 2.3. This experiment evaluates the model's ability to recover token positions, which is a necessary condition for the emergence of spatial structures. See Section C for futher details.

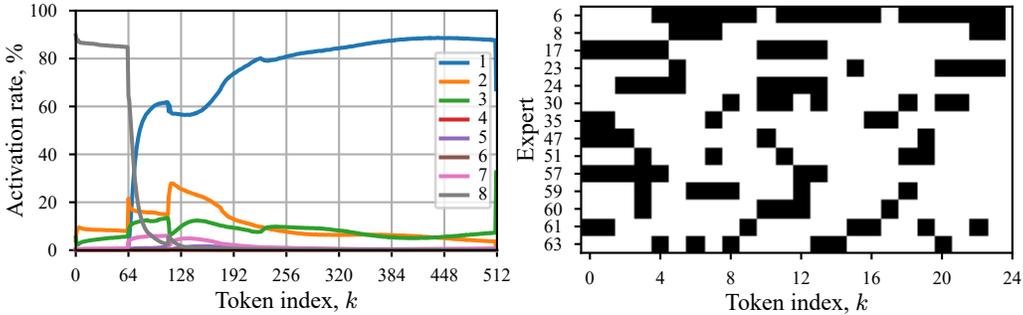

Figure 1: Average expert activation rates in the sixth MoE-block measured on SLIMPAJAMA dataset for SWITCH8 (left) and sample expert activation trajectory for OLMoE (right). The activations trajectory is sampled from the bottom layer for the first 24 tokens with first 14 most active experts.

### 2.1 EXPERT ACTIVATION RATE

We observe two qualitatively distinct behaviours. The first corresponds to SWITCH model (see Figure 1), where certain experts exhibit significantly higher activation rates, forming long sequences of repeatedly active experts with stable rate values. In contrast, the OLMoE model displays a different pattern (see Figure 3), with expert activation rates $r_{ijk}$ fluctuating around an average value $\bar{r}_{ij}$. We hypothesize that the difference arises from the top-1 selection. Alternatively, it could be due to a bug in the implementation provided in TRANSFORMERS (Wolf et al., 2020).

### 2.2 EXPERT CORRELATION LENGTH

Our experiments show that the correlation lengths $\xi_{\text{model}}$ of the models are larger than those of random expert activations, with a mild growth of $\xi$ with depth of MoE-layer $l$, peaking in the middle layers (see Figure 2). This suggests consecutive activations of the same experts, forming patterns influenced by high-level features. Scaling of $\xi_{\text{model}}$ with $N_{\text{block}}$ indicates non-uniform expert activation and decaying long-range correlations. Since MoE-router is independent of token position, we suggest that experts interact with each other through embeddings vectors, which carry positional information from RoPE embeddings within the attention mechanism.

### 2.3 TOKEN POSITION PREDICTION

If MoE-router is capable of capturing positional token information, then a classifier of comparable complexity should be able to capture it as well. While predicting exact token position $k$ is hard merely because of the large number of classes, the classifier should be able to predict simpler synthetic targets, such as parity $2 \mid k$ (even or odd) or the index of the subsequence of $n$ tokens, $\lfloor \frac{k}{n} \rfloor$.

According to Table 1, MoE-router is potentially capable of extracting positional information from embeddings. While it cannot accurately predict the exact token position, it can more reliably predict the blocks index $\lfloor \frac{k}{n} \rfloor$. Interestingly, the most notable difference in predicting is between parity and block index of size 2. The classifier fails to predict parity but can successfully determine whether a token belongs to the first or second half of a sequence. This points out to the importance of low-frequency components of RoPE, while highlighting the limited utility of its high-frequency parts.

## 3 PHENOMENOLOGICAL MODEL

Based on the empirical observations described in Section 2, we propose a phenomenological model in Section 3.1 that characterizes the MoE-model from the perspective of statistical mechanics (SM).

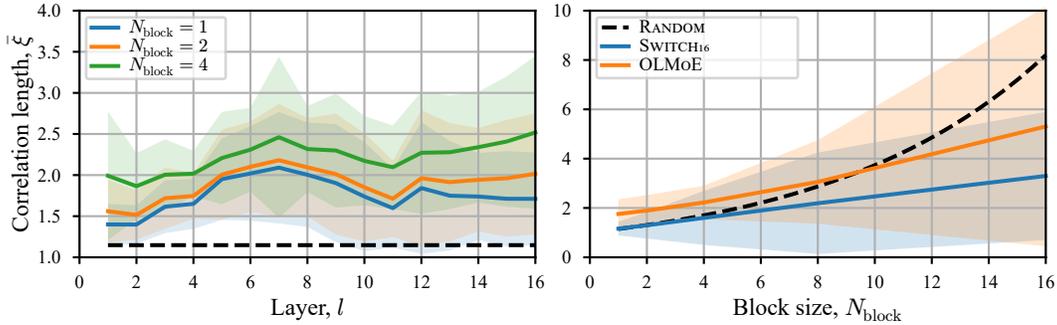

Figure 2: Averaged correlation length $\bar{\xi}$ scaling. Correlation lengths are averaged over experts in OLMoE layers (left) or in the entire model (right).

Subsequently, we address the load balancing problem and suggest a new auxiliary loss within the proposed model in Section 3.2.

### 3.1 STATISTICAL MODEL

Consider a sequence of tokens $t_i$ and corresponding input embeddings $x_i$ with $0 \leq i < L-1$. Each token $t_i$ can be attributed to one of the states $s_i$. Gibbs gives a probability distribution on states

$$p(s_i|\Theta, t_1, t_2, ..., t_{L-1}) \propto e^{-\beta E(s_i;\Theta, t_1, t_2, ..., t_{L-1})}, \qquad (1)$$

where set of all tokens $t_i$ parametrizes energy function $E$ (we omit tokens in the parameter specification from now on) and $\beta = T^{-1}$ is inversed temperature. Natural ordering of tokens and corresponding experts forms a one dimensional lattice (or chain). This chain of experts ( spins in SM) defines a one dimensional Ising model of a mixture of experts governed by Hamiltonian $H(s_i)$.

The form of Hamiltonian $E$ essentially defines all properties of the model. However, it is quite difficult to provide closed analytical expression for $E$ but it is sufficient enough to reason about main characteristics and spin interactions in particular. For example, an attention mechanics ensures non-linear expert-expert interaction but its range is a more subtle topic.

We assume that a typical MoE model like OLMoE (Muennighoff et al., 2024) demonstrates a short-range expert-expert interactions. From general consideration, one may expect that a TRANSFORMER for language modeling tends to operate on a near context of tens or hundreds of tokens. For example, grammar and syntax require agreement and concordance among words in an utterance.

The more specific argument is based on decaying of attention scores. It is a complex and different subject which goes beyond this work and which requires an additional study. However, modern TRANSFORMER-based architectures and models used for empirical study in Section 2 operate internally with a variant of relative positional encodings (Shaw et al., 2018) which admit scores decaying with relative distance between tokens. Specifically, RoPE positional encodings (Su et al., 2022), a building block of OLMoE, do indeed decays (see Section 3.4.3 in (Su et al., 2022)).[1] This constitutes the rationale for using one dimensional model defined above as a phenomenological model of MoE-blocks in TRANSFORMERS.

### 3.2 LOAD BALANCING PROBLEM

The load balancing problem is a problem of assigning equisized batches of tokens to each expert (Fedus et al., 2022a). Without extra efforts, a few most active expert creates a positive feedback loop during training that makes them the only active experts in training and inference time. This well-known fact can be reframed with our spin-glass model.

High activation rates of some experts corresponds high density of these experts (spins) in our spin-glass. Experts can appear in different spatial structures forming topologically ordered phases.

---

[1] More precisely, Su et al. (2022) give only an upper bound on the multiplication factor. Long-range decay of RoPE is also challenged by Barbero et al. (2024).

Table 1: Quality metrics of token position $k$ prediction task against different (syntetic) targets for the first MoE-block in OLMoE. Token position parity is $2 \mid k$. Target for classification on consecutive blocks of size $n$ is denoted as $\lfloor k/n \rfloor$. Shadowed value in parentheses denotes standard deviation.

| Target | Classes | Acc@1 | Acc@2 | Acc@8 | AP | Pr | Recall | F1 |
|---|---|---|---|---|---|---|---|---|
| $2 \mid k$ | 2 | 49.9(5) | — | — | 50.1(7) | 49.9(5) | 50.4(4) | 50.1(4) |
| $\lfloor k/128 \rfloor$ | 2 | 91.7(3) | — | — | 96.4(5) | 90.0(3) | 93.9(3) | 91.9(3) |
| $\lfloor k/64 \rfloor$ | 4 | 75.9(3) | 95.9(2) | — | 82.8(3) | 76.0(3) | 75.9(3) | 75.9(3) |
| $\lfloor k/16 \rfloor$ | 16 | 41.8(1) | 64.7(2) | 96.2(2) | 41.1(4) | 41.3(1) | 41.8(1) | 41.3(1) |
| $\lfloor k/4 \rfloor$ | 32 | 13.8(2) | 24.4(5) | 59.5(6) | 11.8(1) | 13.5(2) | 13.8(2) | 13.6(2) |
| $\lfloor k/2 \rfloor$ | 128 | 6.8(3) | 12.4(5) | 35.2(1) | 5.8(1) | 6.73(2) | 6.8(3) | 6.7(3) |
| $k$ | 256 | 3.6(1) | 6.0(1) | 17.4(3) | 3.0(1) | 3.6(2) | 3.6(1) | 3.5(2) |

However, Landau's argument for absence of ordering in one dimension (Landau and Lifshitz, 2013) breaks any ordering at all.

**Remark 3.1.** In contrast to one dimensional Ising model, the general case of $n > 1$ allows existence of ordering. For example, two dimensional lattices of experts and tokens can emerge from models with augmented context like RETRO (Borgeaud et al., 2022). This particularly means that training of such kind of models could potentially require additional effort in respect to conventional MoE-models. On the other hand, the absence of ordering in one dimension could imply reduced model expressivity since some expert configurations are topologically prohibited.

The only remaining condition that must be met is equilibrium state, i.e. entropy $H(p)$ is maximal.

$$\max_{\Theta} \mathcal{L}_{\text{MEM}}(p; \Theta), \quad \mathcal{L}_{\text{MEM}}(p; \Theta) = T \sum_{i=1}^{L} H(p(s_i)). \quad (2)$$

In machine learning literature, this method is known as maximum-entropy principle (Jaynes, 1957a; 1957b). Obviously, our MEM-loss should be taken with negative sign in order to be used as an auxiliary loss term for end-to-end training of entire model. Optimization problem (2) can be rewritten as a minimization problem of KL-divergence between the proposed distribution and the equilibrium:

$$\min_{\Theta} \ T \sum_{i=1}^{L} D_{\text{KL}}(p(s_i; \Theta) \parallel q). \quad (3)$$

Upper bound of $H(p)$ for discrete states $s_i$ corresponds to uniform distribution on $s_i$ with probability $q_i$. Since there are $\binom{n}{k}$ different states of $k$ active experts out of $n$ experts, $q_i = 1/\binom{n}{k}$. For example, $q = 1/\binom{64}{8} \approx 2.26 \cdot 10^{-10}$ in case of OLMoE.

## 4 Practical Implications

Observations made in Section 2 lead to several practical implications which we discuss here.

**Observation 4.2** *(Spatial structure)* Figure 2 and Table 1 suggest that a spatial correlation among experts exists and experts tend to form spatial structures.

Observation 4.2 motivates study of a static routing as alternative to the dynamic gating mechanism in MoE. Static routing means an expert activation in dependence on its position. It has multiple potential advantages in comparison to dynamic gating. First, static routing is less sensitive to data shuffling and uneven communications among model shards with `scatter` and `gather` primitives. Second, training complications like load balancing heuristics and loss terms can be neglected.

**Observation 4.3** Statistical mechanics suggests a valid phenomenological model of MoE.

Observation 4.3 supports the applicability of the entropy maximization principle (2). This results in the formulation of the MEM-loss (3) as an alternative to aux-loss (Fedus et al., 2022a) or Z-loss (Zoph et al., 2022). MEM-loss offers a clear interpretation and is theoretically grounded. However, its practical validation in training is left for future work.

## 5 COMMENTS AND DISCUSSION

In this work, we studied internal MoE dynamics empirically. Specifically, we found and experimentally demonstrated spatial correlations in expert activations. Token position prediction experiment highlights importance of positional information for entire TRANSFORMER architecture. The phenomenological model provides a perspective of statical mechanics and motivates MEM-loss, a theoretically grounded alternative to load balancing loss. Training MoE-model from scratch with MEM-loss and experimenting with bigger models and models of different architectures are left for future work.

## A  Related Works

Comprehensive surveys of Mixture of Experts (MoE) can be found in (Yuksel et al., 2012) and (Fedus et al., 2022b). The first work (Yuksel et al., 2012) focuses on early works in the first twenty years of MoE development and the second one (Fedus et al., 2022b) cover the next ten years. Initially, MoE approach has been described in (Jacobs et al., 1991) then it has been generalized to broaded class of hierarchical models in (Jordan and Jacobs, 1994).

**Combinatorical properties.** In work (Jiang, 2000) authors studed combinatorial properties of MoE models. Specifically, they provided bounds for mixture of Bernoulli classifier and mixture of logistic regression classifier.

**Statistical properties.** In work (Kang and Oh, 1996) authors considered MoE models from statistical mechanics.

## B  Available MoE-Transformers

In all experiments, we use only pretrained models published on HuggingFace Hub. We make a long list of recent Transformers models with MoE-adapter presented in literature (Dai et al., 2024; Du et al., 2022; Jiang et al., 2024; Lepikhin et al., 2020; Muennighoff et al., 2024; Rajbhandari et al., 2022; Xue et al., 2024; Zhu et al., 2024; Zoph et al., 2022) (see Table 2). Despite the fact there are plenty of models available online, we are limited by two factors. Firstly, only a few models have a necessary instrumentation required to get MoE router output logits and selected experts. The second limitation stems from hardware available to us (i.e. 2x Nvidia V100). It's getting even worse with the fact that the majority of models use bfloat16 for arithmetics which has no native support on GPUs of Volta family (fortunately, emilation via fp32 rescues the day).

We use PyTorch for our experiments. For model inference, autodiff is disable with `torch.inference_mode()` context but models stay in training mode (i.e `.training` attribute evaluates to true). In this way, we save memory for larger batch and sample expert activations as if we indeed train a model.

## C  Miscellaneous Empirical Study

In order to ensure versatility among model architectures as well as MoE layer architectures, we consider MoE-models of distinct two architectures: Switch (Fedus et al., 2022a) and OLMoE (Muennighoff et al., 2024). Switch is an encoder-decoder Transformer based on T5 (Raffel et al., 2020) with a single active expert and expert capacity $C_{\text{expert}} = 64$. We use its two variants with total number of experts $N_{\text{expert}} = 8$ and $N_{\text{expert}} = 16$ (Switch8 and Switch16). Despite the encoder-decoder nature of Switch, we pass an empty premise to encoder. OLMoE is a decoder-only transformer with 8 active experts out of total $N_{\text{expert}} = 64$ experts. Pre-trained model checkpoints published on the HuggingFace Hub were used in all experiments (see Section B). In addition, we use a test split of a diverse SlimPajama corpus (Soboleva et al., 2023) to guarantee the representativeness samples of expert activations.

### C.1  Expert Activation Rate

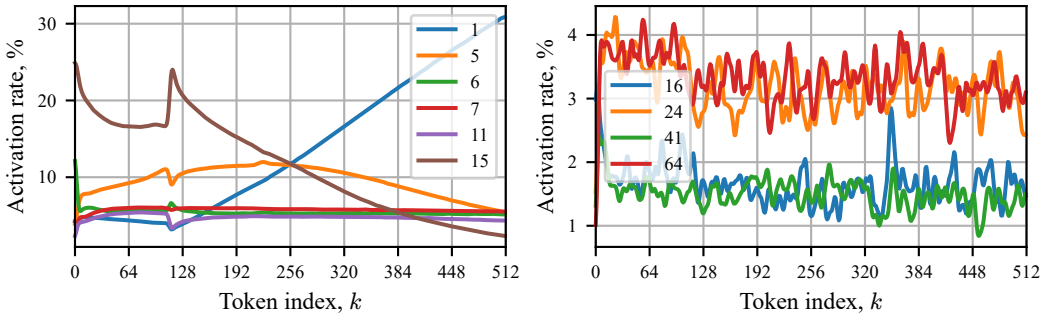

Figure 3: Expert activation rates for the first 512 tokens and some experts of SWITCH16 (left) and OLMoE (right) models. Expert choice is purely random with a minor exception. Experts 1 and 15 of SWITCH16 are outliers and demonstrates atypical behavior. Expert activation rates of OLMoE are smoothed for better representation with Gaussian filter of $\sigma = 2$.

In order to measure in what degree experts are interleaved or overlapped each other, we collect expert activation frequencies for specific token position and TRANSFORMER-block on a sample sequences. Sequence are sampled from test split of SLIMPAJAMA corpus with about 500k documents. Then aggregated frequencies over all samples $c_{ijk}$ are used to estimate expert activation rates as $r_{ijk} = c_{ijk}/\sum_k c_{ijk}$ (see Figure 1 and 3).

### C.2 EXPERT CORRELATION LENGTH

Table 2: Incomplete list of availabel MoE models that can potentially be used in for experimentation.

| MODEL | REFERENCE | PARAMETERS | HUGGINGFACE (🤗) |
|---|---|---|---|
| GShard | (Lepikhin et al., 2020) | 37B | — |
| DeepSpeed-MoE | (Rajbhandari et al., 2022) | 350M/13B | — |
| | | PR-350M/4B | — |
| | | PR-1.3B/31B | — |
| GLAM | (Du et al., 2022) | 0.1B/1.9B | — |
| | | 1.7B/27B | — |
| ST-MoE | (Zoph et al., 2022) | 0.8B/4.1B | — |
| Switch | (Fedus et al., 2022a) | 250M | google/switch-base-8 |
| | | 1B | google/switch-base-16 |
| | | 2B | google/switch-base-32 |
| | | 4B | google/switch-base-64 |
| DeepSeek-MoE | (Dai et al., 2024) | 0.24B/1.89B | — |
| | | 2.8B/16.4B | deepseek-ai/deepseek-moe-16b-base |
| LLaMA-MoE | (Zhu et al., 2024) | 3.0B/6.7B | llama-moe/LLaMA-MoE-v1-3_0B-2_16 |
| | | 3.5B/6.7B | llama-moe/LLaMA-MoE-v1-3_5B-4_16 |
| | | 3.5B/6.7B | llama-moe/LLaMA-MoE-v1-3_5B-2_8 |
| Mixtral | (Jiang et al., 2024) | 13B/47B | mistralai/Mixtral-8x7B-v0.1 |
| | | 39B/141B | mistralai/Mixtral-8x22B-v0.1 |
| OLMoE | (Muennighoff et al., 2024) | 1B/7B | allenai/OLMoE-1B-7B-0924 |
| OpenMoE | (Xue et al., 2024) | 8B | OrionZheng/openmoe-8b |

Table 3: Fitted parameters of law $\xi/\xi_0 = \exp(\alpha n)$ to experimental data.

| Model | $N_{\text{expert}}$ | $\alpha$ | $\xi_0$ |
|---|---|---|---|
| Random | 1/8 | 0.131269 | 0.005969 |
|  | 1/16 | 0.131088 | 0.006606 |
| OLMoE | 8/64 | 0.073877 | 0.499299 |
| Switch16 | 1/16 | 0.068579 | 0.146133 |

We use two definitions of correlation length. The first one $\xi_{\text{dw}}$ is defined as the ratio of the sequence length $L$ to the average number of domain walls $\bar{N}$. The second one $\xi_{\text{ds}}$ is estimated with direct computation of domain sizes $L_e$ with subsequent averaging.

$$\xi_{\text{dw}} = \frac{L}{\bar{N}_e}, \quad \xi_{\text{ds}} = \bar{L}_e. \tag{4}$$

In practice, $\xi_{\text{dw}}$ and $\xi_{\text{ds}}$ are correlated. However, $\xi_{\text{ds}}$ is more precise definition but more costly to estimate than $\xi_{\text{dw}}$. Henceforth, $\xi$ is used to denote $\xi_{\text{ds}}$ without label.

Domain size is estimated over block variables, i.e. block of $N_{\text{block}}$ sequential expert activation indicators. If an expert $k$ is activated in merely one indicator of a block, than the entire block indicates activation of expert $k$ (i.e. disjunctive union of experts).

From analysis of experimental data, we find that $\xi$ scales well with $n$ according exponential law, i.e.

$$\frac{\xi}{\xi_0} = e^{\alpha n}. \tag{5}$$

We fit exponents $\alpha$ for models of interest and present them in Table 3.

### C.3 Token Position Prediction

We dissect OLMoE and keep only the first layer up to MoE-router. We sample embeddings short sequences of length $L = 256$ right before MoE-router but after layer normalization (normalization is critical for training a linear models). Finally, we train a multinomial logistic with Scikit-Learn (Pedregosa et al., 2011) with stratified 3-fold cross validataion and grid search over $L_2$ reguralizer.